\documentclass{article}

\usepackage{arxiv}

\usepackage[utf8]{inputenc} 
\usepackage[OT2,T1]{fontenc}    
\usepackage{hyperref}       
\usepackage{url}            
\usepackage{booktabs}       
\usepackage{amsfonts}       
\usepackage{nicefrac}       
\usepackage{microtype}      
\usepackage{graphicx}
\usepackage{natbib}
\usepackage{doi}
\usepackage[russian,english]{babel}

\usepackage{times}
\usepackage{latexsym}
\usepackage{algorithm}
\usepackage{latexsym}
\usepackage{algpseudocode}
\usepackage{multicol}
\usepackage{amsmath}
\usepackage{enumitem}
\usepackage{multirow} 
\usepackage{algorithmicx}
\usepackage{tipa}

\usepackage[T1]{fontenc}

\usepackage{microtype}

\usepackage{inconsolata}

\title{Colexifications for Bootstrapping Cross-lingual Datasets: The Case of Phonology, Concreteness, and Affectiveness}

\author{Yiyi Chen\thanks{This work is supported by the Carlsberg Foundation under a \textit{Semper Ardens: Accelerate} career grant held by JB, entitled ``Multilingual Modelling for Resource-Poor Languages'', grant code CF21- 0454. This work is accepted to SIGMORPHON 2023. } \\
	Department of Computer Science\\
	Aalborg University\\
	Copenhagen, Denmark \\
	\texttt{yiyic@cs.aau.dk} \\
	\And
	Johannes Bjerva \\
	Department of Computer Science\\
	Aalborg University\\
	Copenhagen, Denmark \\
	\texttt{jbjerva@cs.aau.dk} 
}

\renewcommand{\shorttitle}{Colexifications for Bootstrapping Cross-lingual Datasets}

\begin{document}
\maketitle

\begin{abstract}
Colexification refers to the linguistic phenomenon where a single lexical form is used to convey multiple meanings. 
By studying cross-lingual colexifications, researchers have gained valuable insights into fields such as psycholinguistics and cognitive sciences~\citep{jackson-2019,XU2020104280,karjus2021conceptual,SchapperKoptjevskajaTamm+2022+199+209,François+2022+89+123}.
While several multilingual colexification datasets exist, there is untapped potential in using this information to bootstrap datasets across such semantic features.
In this paper, we aim to demonstrate how colexifications can be leveraged to create such cross-lingual datasets. We showcase curation procedures which result in a dataset covering 142 languages across 21 language families across the world.
The dataset includes ratings of concreteness and affectiveness, mapped with phonemes and phonological features.
We further analyze the dataset along different dimensions to demonstrate potential of the proposed procedures in facilitating further interdisciplinary research in psychology, cognitive science, and multilingual natural language processing (NLP). 
Based on initial investigations, we observe that i) colexifications that are closer in concreteness/affectiveness are more likely to colexify; ii) certain initial/last phonemes are significantly correlated with concreteness/affectiveness intra language families, such as \textipa{/k/} as the initial phoneme in both Turkic and Tai-Kadai correlated with concreteness, and \textipa{/p/} in Dravidian and Sino-Tibetan correlated with Valence; 
iii) the type-to-token ratio (TTR) of phonemes are positively correlated with concreteness across several language families, while the length of phoneme segments are negatively correlated with concreteness; iv) certain phonological features are negatively correlated with concreteness across languages. 
The dataset is made public online for further research\footnote{\url{https://github.com/siebeniris/ColexPhon}}.
\\

\end{abstract}


\section{Introduction}

Semantic typology studies cross-lingual semantic categorization~\citep{evans2010semantic}. 
Within this area, the term ``colexification" was first introduced and used by~\citet{franccois2008semantic} and~\citet{haspelmath2003geometry} to create semantic maps. The study of colexifications focuses on cross-lingual colexification patterns, where the same lexical form is used in distinct languages to express multiple concepts. For instance, \textit{mapu} in Mapudungun and \textit{apakee} in Ignaciano both express the concepts \textsc{earth} and \textsc{world}~\citep{clics}. 
Colexifications have been found to be pervasive across languages and cultures. The investigation of colexifications have led to interesting findings across different fields, such as linguistic typology~\citep{SchapperKoptjevskajaTamm+2022+199+209}, psycholinguistics~\citep{jackson-2019}, cognitive science~\citep{GIBSON2019389}, but remain relatively unexplored in NLP~\citep{harvill-etal-2022-syn2vec,chen2023colex2lang}.

\begin{figure}[tp]
    \centering
    \includegraphics[width=0.6\textwidth]{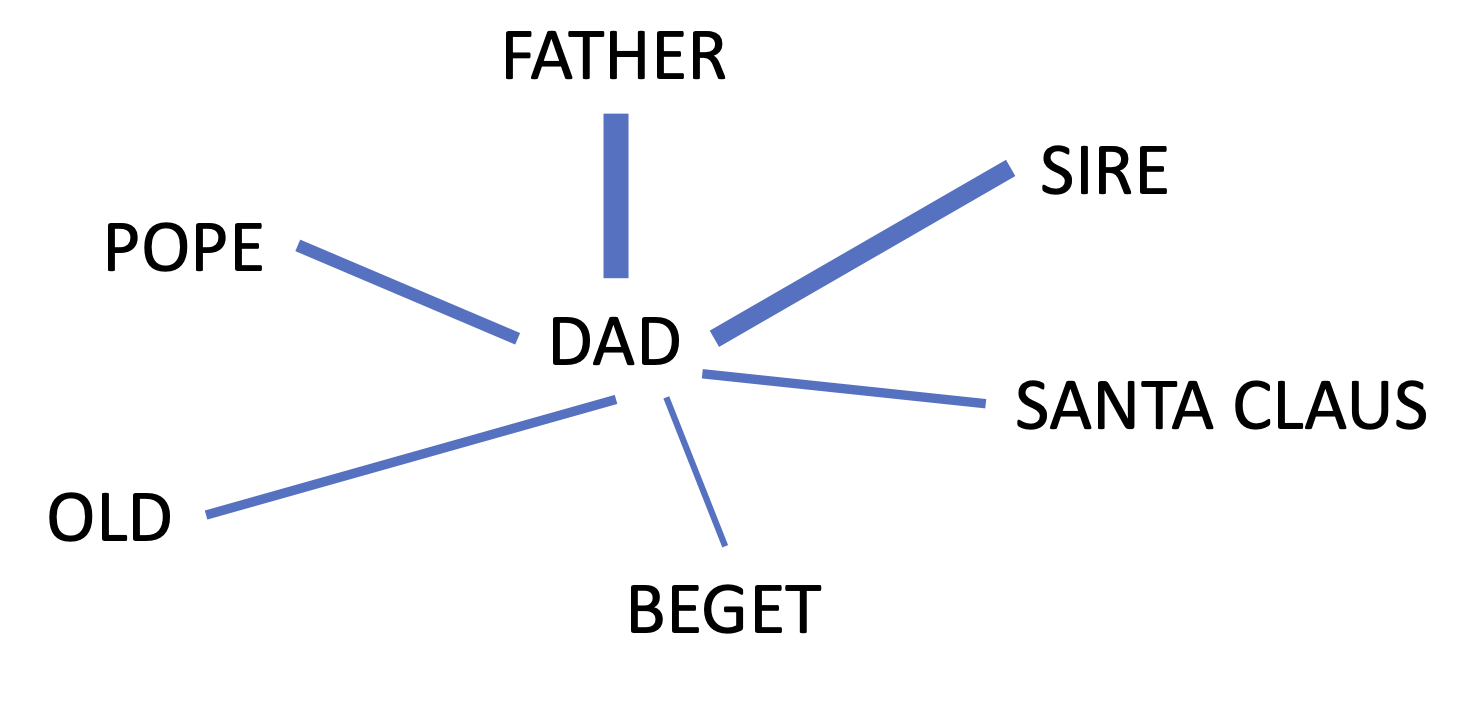}
    \caption{Colexification subgraph for \textsc{DAD}. The weight of the edges are proportional to the frequency of the colexification pattern in the dataset. }
    \label{fig:colex}
\end{figure}

In recent years, with the increasing popularity of automatic methods and big data in linguistics, datasets such as Concepticon~\citep{clld2022} and BabelNet~\citep{babelnet} have been developed, affording large-scale cross-lingual semantic comparisons.
The Database of Cross-lingual Colexifications (CLICS\textsuperscript{3})~\citep{clics} was created based on the Concepticon concepts, including 4,228 colexification patterns across 3,156 languages, to facilitate research in colexifications. Studies have also been shown to curate large-scale colexification networks from BabelNet, consisting of over 6 million synsets across 520 languages~\citep{harvill-etal-2022-syn2vec, chen2023colex2lang}.

While syntactic typology is relatively well-established in NLP \citep{malaviya-etal-2017-learning,bjerva2018phonology,bjerva2018tracking,bjerva-augenstein-2021-typological,cotterell-etal-2019-complexity,bjerva-etal-2019-probabilistic,bjerva-etal-2019-uncovering,bjerva-etal-2019-language,bjerva2020sigtyp,stanczak-etal-2022-neurons,ostling-kerfali-2023,fekete-bjerva-2023-gradual}, semantic typology has so far only been subject to limited research \citep{chen2023colex2lang,chen2023americas,liu2023crosslingual}.
As a relatively new topic in both semantic typology and NLP, colexifications covers a wide-range of languages and language families. 
In contrast, although the concepts of concreteness/abstractness and affectiveness (e.g., valence, dominance and arousal) have long been in the center stage of interdisciplinary research fields such as cognitive science, psychology, linguistics and neurophysiology~\citep{warriner2013norms,solovyev2021concreteness,brysbaert2014concreteness}, language coverage of such resources is severely limited, and curation prohibitively expensive.

The study of phonemes and phonological features have furthermore been essential to, e.g., address the problems of non-arbitrariness in languages and investigating universals of spoken languages~\citep{https://doi.org/10.1111/cogs.13147}. 
Studies such as~\citet{GastKoptjevskajaTamm+2022+403+438} demonstrate the genealogical stability (persistence) and susceptibility to change (diffusibility) via studying the patterns the phonemes/phonological forms and the colexifications across European languages. 
However, this study is limited to a small range of languages, and the investigated concepts are also restricted to 100-item Swadesh list~\citep{swadesh1950salish}. With the proposed procedures, a wider range of concepts and the phonological forms across language families are curated.

In this paper, we create a synset graph based on multilingual WordNet~\citep{wordnet} data from BabelNet 5.0. 
We then develop a cross-lingual dataset that includes ratings of concreteness and affectiveness, as this approach yields more comprehensive data than using CLICS\textsuperscript{3}. 
In addition, we meticulously select and organize phonemes and phonological features for the lexicons that represent the concepts. 
Our methodology for data creation is not limited to the constructed dataset, as it has potential for broader applications. 
We showcase the versatility of our approach through analysis across various dimensions, and make our dataset freely available.

\section{Related Work}

\paragraph{Colexifications}

The creation of semantic maps using cross-linguistic colexifications was initially formalized by~\citet{franccois2008semantic}. Semantic maps are graphical representations of the relationship between recurring expressions of meaning in a language~\citep{haspelmath2003geometry}. This method is based on the idea that language-specific colexification patterns indicate the semantic proximity or relatedness between the meanings that are colexified~\citep{hartmann2014identifying}. When analyzed cross-linguistically, colexification patterns can provide insights into various fields, such as cognitive principles recognition~\citep{berlin1991basic,SchapperRoqueHendery+2016+355+422,jackson-2019,GIBSON2019389,XU2020104280,BROCHHAGEN2022105179}, diachronic semantic shifts in individual languages~\citep{Witkowski1985,urban2011,karjus2021conceptual,François+2022+89+123}, and language contact evolution~\citep{heine_kuteva2003,koptjevskaja-tamm_liljegren_2017,SchapperKoptjevskajaTamm+2022+199+209}.

\citet{jackson-2019} conducted a study on cross-lingual colexifications related to emotions and found that different languages associate emotional concepts differently. For example, Persian speakers associate \textsc{grief} closely with \textsc{regret}, while Dargwa speakers associate it with \textsc{anxiety}. The variations in cultural background and universal structure in emotion semantics provide interesting insights into the field of NLP.~\citet{bao-etal-2021-universal} analyzed colexifications from various sources, including BabelNet, Open Multilingual WordNet, and CLICS\textsuperscript{3}, and demonstrated that there is no universal colexification pattern.

In the field of NLP, ~\citet{harvill-etal-2022-syn2vec} constructed a synset graph from BabelNet to boost performance on lexical semantic similarity task. More recently,~\citet{chen2023colex2lang} use colexifications to construct language embeddings and further model language similarities. 
Our goal is to utilize colexifications to construct cross-lingual datasets, including diverse ratings and phonological forms and features, to support further research, particularly in low-resource languages where norms and ratings are notably scarce.

\paragraph{Norms and Ratings}
A large number of words in high-resource languages have been assigned norms and ratings by researchers in psychology~\citep{brysbaert2014concreteness, warriner2013norms}. 
Norms and ratings of words are essential components in psychology, linguistics, and recently being widely used in NLP. Norms refer to the typical frequency and context in which words are used in a particular language, while ratings represent subjective judgements of individuals on various dimensions such as concreteness, valence, arousal, and imageability. 
These norms and ratings can improve the performance on downstream tasks, such as sentiment analysis, emotion recognition, word sense disambiguation, and affective computing~\citep{kwong-2008-preliminary,tjuka2022linking,strapparava-mihalcea-2007-semeval, mohammad-turney-2010-emotions}.

The study of concreteness and abstractness of concepts is interdisciplinary and spans across various fields, including linguistics, psychology, psycholinguistics, and neurophysiology~\citep{solovyev2021concreteness}. Concrete concepts are those that can be perceived by the senses, such as \textsc{cat} and \textsc{mountain}, while abstract concepts, like \textsc{relationship} and \textsc{understanding}, cannot be perceived by the senses.~\citet{brysbaert2014concreteness} conducted a study on concreteness ratings for 37,058 English words and 2,896 two-word expressions, involving over 4,000 participants, which has provided insights across various linguistic disciplines. The concreteness ratings are based on a scale of 1 (abstract) to 5 (concrete).
These ratings have been used in conjunction with various tasks such as classification of metaphoricity \citep{haagsma2016detecting} and animacy \citep{bjerva2014multi}, as well as cultural studies \citep{berger2022using}.

Apart from concreteness, affective ratings are also essential for interdisciplinary research in psychology, linguistics and NLP. The affective norms for English words (ANEW) dataset, providing ratings of valence, arousal and dominance for English words, has been widely used in both psychology and NLP research~\citep{Bradley1999AffectiveNF}. Subsequently, the affective norms for French Words (FAN) and the affective norms for German words (ANGST) datasets, proving similar affective ratings for French and German words, respectively, have also been developed~\citep{monnier2014affective,schmidtke2014angst}. The Spanish version of ANEW is developed by~\citet{redondo2007spanish}. Extending the English ANEW,~\citet{warriner2013norms} covers nearly 14,000 English lemmas, providing ratings for valence (the pleasantness of a stimulus), arousal (the intensity of emotion provoked by a stimulus), and dominance (the degree of control exerted by a stimulus). For creating our dataset, we use the ratings from~\citet{warriner2013norms}, see details in Section~\ref{sec:procedures}.

The data for linguistic norms and ratings is usually collected only for one language. For low-resource languages, such data is obviously lacking. Using our procedures, the norms and ratings can be bootstrapped for low-resource languages by sharing cross-lingual concepts through colexifications.

\paragraph{Phonemes and Phonological Features}
While direct phonetic comparison across languages is difficult, a common practice in comparing phonological characteristics across languages is to combine similar sounds into one multilingual phone set \citep{salesky-etal-2020-corpus}.
While more advanced methods for phonological typology do exist, e.g.~\citet{cotterell-eisner-2017-probabilistic,cotterell-eisner-2018-deep}, a basic approach to phonology is found via the International Phonetic Alphabet (IPA), which classifies sounds based on general phonological properties.
In this vein, WikiPron is created to serve as an open-source tool for mining phonemic pronunciation data from Wikitionary and still under continuous maintenance~\citep{lee-etal-2020-massively}. To this date, it contains more than 1,8 million word/pronunciations across 543 languages.\footnote{\url{https://github.com/CUNY-CL/wikipron}}
The pronunciations are given in IPA, and segmented in a way that IPA diacritics can be properly recognized~\citep{lee-etal-2020-massively}.

Demonstrating that phonological features outperform character-based models, PanPhon is created and used for various NER-related tasks~\citep{Mortensen-et-al:2016}. To date, PanPhon is a database relating over 5,000 IPA segments to 24 subsegmental articulatory features.\footnote{\url{https://github.com/dmort27/panphon}}
It has been used for various purposes, such as cross-modal and cross-lingual study of iconicity in languages~\citep{zhu2021multilingual}, and cross-linguistic phonosemantic correspondence using a deep-learning framework~\citep{de2021layered}.

In this paper, we build upon this work by diving into the relationship between phonological features, and the concreteness and affectiveness of sense lemmas across a wide set of languages.
The paper is inspired by findings such that the sounds of words can influence their meaning and emotional impact. 
For example, words with round vowel sounds are often associated with positive emotions, while harsher, more angular sounds can convey negative emotions~\citep{cwiek2022bouba}. 
This study aims to initiate the study on the intricate interplay between sound and affective/abstract meanings.

\section{Dataset Curation}\label{sec:procedures}

A \textit{colexification pattern} refers to a case where two concepts are colexified, such as \textsc{dad}-\textsc{pope} shown in Figure~\ref{fig:colex}. Specifically, a \textit{colexification} is an instance of a \textit{colexification pattern}, such as \textit{far} in Danish, as shown in Table~\ref{tab:example}.

\begin{figure}[htb]
    \centering
    \includegraphics[width=0.6\textwidth]{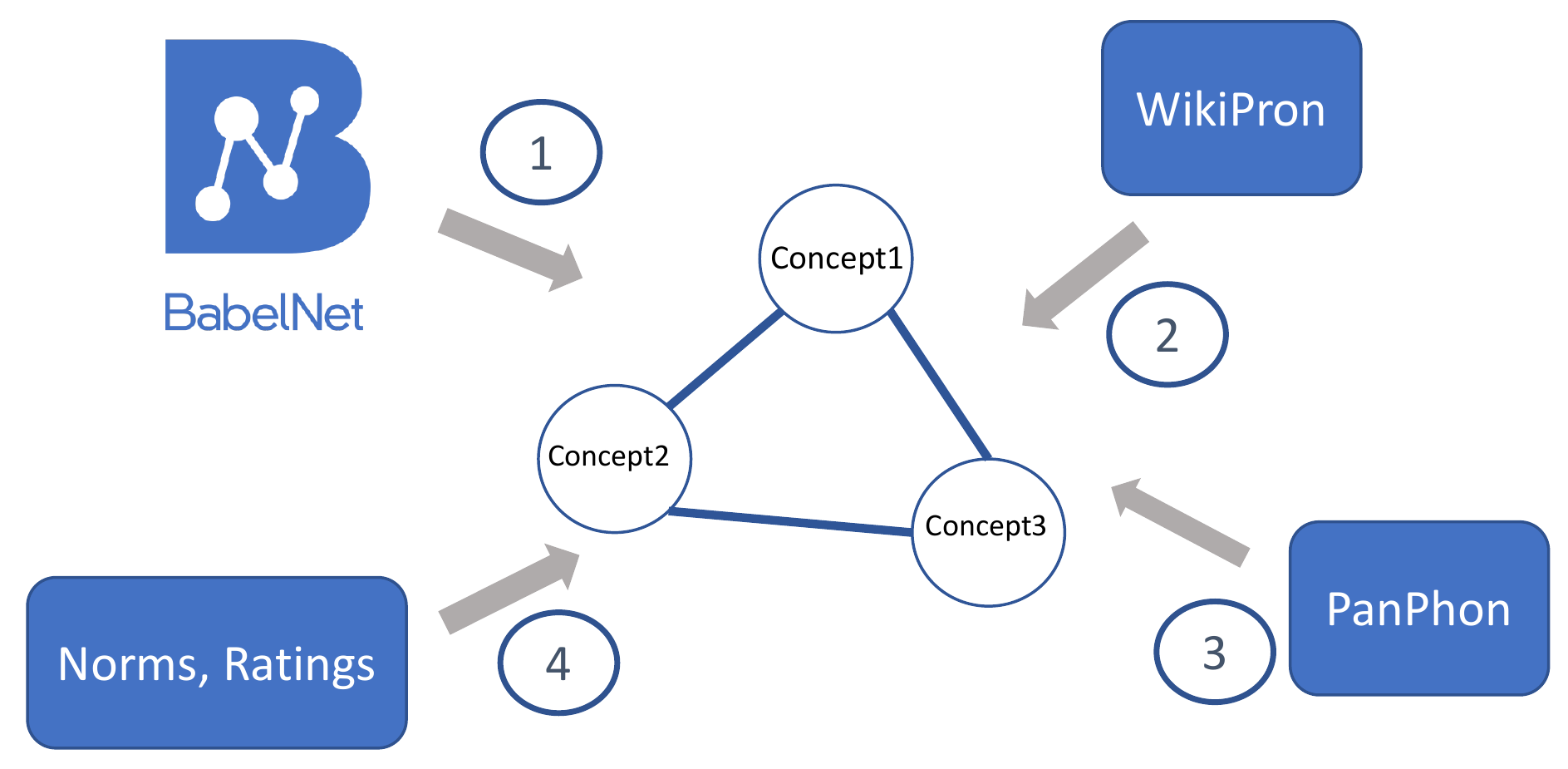}
    \caption{The Workflow of the Procedures for Creating the cross-lingual Dataset using Colexifications. }
    \label{fig:workflow}
\end{figure}

In order to leverage colexifications to create a cross-lingual dataset incorporating norms and ratings in psychology and other fields, we propose the following procedures for data curation and creation, as illustrated in Fig.~\ref{fig:workflow}. 

\begin{table*}[]
    \centering
     \resizebox{\textwidth}{!}{
    \begin{tabular}{ccccccccccc}
    \hline
   
      \textbf{Sense Lemma}   &  \textbf{Language} & \textbf{Phonemes} &  \textbf{Synset 1} & \textbf{Synset 2}  &  \textbf{Concept 1} & \textbf{Concept 2} & \textbf{Conc.Dist} & \textbf{V.Dist} & \textbf{A.Dist} & \textbf{D.Dist}  \\\hline
     \textsc{pāp} & Persian & \textipa{ p A: p} & dad\#n\#1 & pope\#n\#1 & \textsc{dad} & \textsc{pope} & 0.42 & 1.96 & 0.16 & 1.88 \\
      \textsc{bābā} & Arabic & \textipa{ b A: b A:} & dad\#n\#1 & pope\#n\#1 & \textsc{dad} & \textsc{pope} & 0.42 & 1.96 & 0.16 & 1.88 \\
      {\sffamily\foreignlanguage{russian}{papa}} & Russian & \textipa{p a p @}  & dad\#n\#1 & pope\#n\#1 & \textsc{dad} & \textsc{pope} & 0.42 & 1.96 & 0.16 & 1.88 \\
    far & Danish & -  & dad\#n\#1 & sire\#n\#1 & \textsc{dad} & \textsc{sire} & - & 0.74 & 0.05 & 0.57 \\ 
      pare & Castilian & \textipa{p a R e} & Santa\_Claus\#n\#1 & dad\#n\#1 & \textsc{santa claus} & \textsc{dad} & 0.17 & - & - & - \\\hline
     
    \end{tabular}}
    \caption{An example of the dataset. \{CONC,V,D,A\}.Dist represent the distance of the concreteness, valence, dominance and arousal of the pair of concepts for each lexicon. The value is unknown(-) if either of the concepts does not have a rating. }
    \label{tab:example}
\end{table*}

\paragraph{Building the Synset/Concept Graph} In WordNet, a sense is a discrete representation of one aspect of the meaning of a word. For example, the lemma \textit{bank} can either mean the sense \textsc{financial institution} or the sense \textsc{sloping mound}. The set of near-synonyms for a sense is called a \textbf{synset}, which is a primitive in WordNet~\citep{jurafsky2019speech}. Synsets are groups of words sharing the same concept. In order to construct of colexification networks, i) the WordDNet synsets are extracted from BabelNet; ii) for each synset, all the included word senses with their lemmas in the regarding language are elicitated; iii) finally, the sets of synsets sharing the same lemmas are extracted to represent a sysnet graph, with nodes being the synsets and the edges being the lemmas and their languages. The construction of a synset graph from BabelNet is first formalized in~\citep{harvill-etal-2022-syn2vec}, and adapted by~\citep{chen2023colex2lang} incorporating information of the languages and lemmas, see the Algorithm~\ref{algo:cg}.

We adopt the algorithm presented in~\citet{chen2023colex2lang} to construct a large-scale synset graph from WordNet synsets for our study. The difference in~\citet{chen2023colex2lang} and~\citet{harvill-etal-2022-syn2vec} lies in the addition of $G_s$ at line 3 and line 9, as shown in Algorithm~\ref{algo:cg}. $G_s$ affords the construction of colexification patterns and modeling language relations.


\begin{algorithm}[H]
\caption{Construction of Colexification Graph: Given a set of languages L and corresponding vocabularies V, create graph edges between all colexified synset pairs (nodes), consisting of the set of tuples of lemmas and their language.}\label{algo:cg}

\begin{algorithmic}[1]
\Function{CONSTRUCTGRAPH}{$L$,$V$}

  \State $CSP \gets \{\}$    \Comment{Colexified Synset Pairs}
   \State $G_s \gets \mathbf{graph}$ 
    \For{$l\in L$}
        \For{$x \in V_{l}$ }
            \If{$|S_x| \geq 2$}
                \For{$\{s_1,s_2\}\in \binom{S_x}{2}$ } 
                  \State $CSP \gets CSP\cup \{s_i,s_j\}$
                   \State $G_s(s_1, s_2) \gets \{x,l\}$
                \EndFor
            \EndIf
        \EndFor
    \EndFor
    \State $G \gets \mathbf{graph}$ 
    \For{${s_1, s_2} \in CSP$} 
        \State $G(s_1,s_2) \gets 1$
    \EndFor
    \State \textbf{return} $G$
    \State \textbf{return} $G_s$          
\EndFunction

\end{algorithmic}
\end{algorithm}

A WordNet synset comprises a sense word, a Part-of-speech (POS) tag, and a sense number, e.g., \texttt{dad\#n\#1}. The sense numbers indicate the prevalence of the use of senses, with the most frequently used sense labeled 1. The frequency of use is determined by how often a sense is tagged in semantic concordance texts.\footnote{\url{https://wordnet.princeton.edu/documentation/wndb5wn}}
Our assumption is that the mean score of lexicon ratings, annotated by multiple  humans across domains and languages, represents the ratings for the most prevalent sense. However, when it comes to cross-lingual synset-to-concept mapping, there may be variations in the sense annotations between languages.  Suppose that in French the main sense \textsc{knot} is \texttt{knot\#n\#4}, which refers to \textit{a unit of speed}, while in English, the annotation for \textsc{knot} likely refers to \textit{an actual knot that you tie}, which is the 1st sense for the synset. As a result, we cannot expect the same ratings of concreteness or affectiveness for these two different senses. Therefore, to map synsets to concepts, we always select the initial sense of the synsets.. 

Once filtered by the 1st sense of the synsets, as illustrated in Table~\ref{tab:example}, we derive concepts by extracting the sense word from each synset. The resulting concept graph comprises nodes representing the 1st senses of synsets and edges indicating the corresponding languages and sense lemmas.

\paragraph{Phonemes Extraction}
To facilitate analysis of phonetic characteristics cross-lingually in the context of colexifications and against ratings of concreteness and affectiveness, we extract phonemes from WikiPron, which to this date includes 1,882,240 word/pronunciation pairs in 543 languages.\footnote{\url{https://github.com/CUNY-CL/wikipron}}
To map the pronunciations to our data, we mapped their word/language code pairs to the pairs of sense lemma/language code extracted from BabelNet. As a result, there are 139,698 sense lemma/ phonemes pairs across 142 languages, presented as in Table~\ref{tab:example}. In our dataset, the median size of the phonemes per language is 32.

\paragraph{Phonological Features Extraction}
Phonological features have been proposed as the foundation of spoken language universals. Despite variations in phones across languages, the set of phonological features remains constant. Phones can be constructed from a set of phonological features. In our study, we extract phonemes for sense lemmas and then further extract phonological (articulatory) features based on the subsegments using PanPhon. PanPhon generates 24 phonological features for each segment, such as syllabic, sonorant, consonantal, continuant, delayed release, lateral, nasal, strident, voice, spread glottis, constricted glottis, anterior, coronal, distributed, labial, high (vowel/consonant, not tone), low (vowel/consonant, not tone), back, round, elaric airstream mechanism (click), tense, long, hitone, hireg~\footnote{\url{https://github.com/dmort27/panphon}}. Each feature is assigned a value of `1', `-1', or '0', where '1' indicates a positive value of the feature, '-1' indicates a negative value of the feature, and '0' indicates that the feature is absent for that sound. For instance, a vowel cannot possess consonant features, so it is marked as `0'. We use PanPhon to convert each phone into a vector with length 24 in our dataset.

\begin{figure}[htb]
    \centering
    \includegraphics[width=0.7\textwidth]{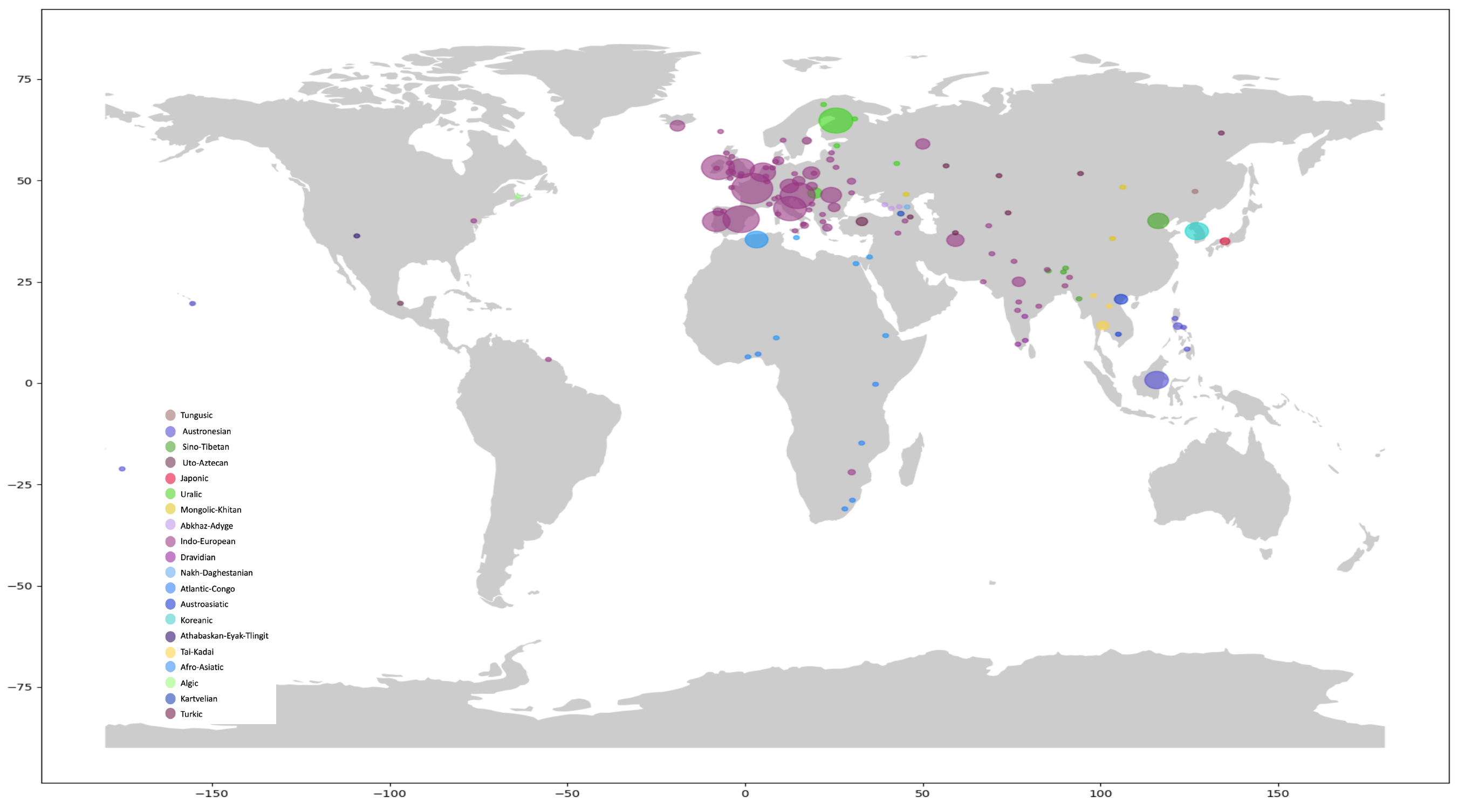}
    \caption{The map of language families of our data. The size of the points are proportional to the number of concepts in each language. Colors represent language families. }
    \label{fig:map}
\end{figure}

\begin{table*}[htb]
    \centering
    \resizebox{\textwidth}{!}{
    \begin{tabular}{cccccccc}
    \hline 
       \textbf{\#Entries}  & \textbf{Colex. Patterns} & \textbf{\#Synset} & \textbf{\#Lexicalization} & \textbf{\#Phone/Lemma pairs} & \textbf{\#Concept} & \textbf{\#Concept w/ Aff.} & \textbf{\#Concept w/ Conc.} \ \\\hline 
       186,6558  & 676,594 &  72,604 & 68,249 & 613,906& 84,084 & 10,353 & 19,179\\ \hline 
    \end{tabular}}
    \caption{Statistics of the Dataset.}
    \label{tab:dataset}
\end{table*}

\paragraph{Incorporating Norms and Ratings}

Having built the concept graph from the synset graph by selecting the 1st senses of the synsets across languages, we map the concepts from databases containing norms and ratings to the concept graph. As shown in Table~\ref{tab:example}, the concept 1 \textsc{DAD} is mapped from concreteness/affectiveness rating lists to the synset 1 \texttt{dad\#n\#1}, while the concept 2 \textsc{POPE} is mapped to the synset 2 \texttt{pope\#n\#1} by intersecting the datasets by the sense words. When each concept in the colexification pair has a rating, the distance of the concreteness/affectiveness can be calculated by computing the absolute distance of the two. When concept 1 has a (mean) concreteness of $conc_1$ and concept 2 has a (mean) concreteness of $conc_2$, then the $Conc.Dist$ is calculated as $|conc_1-conc_2|$. Similar procedures are used for computing distance of valence ($V.Dist$), arousal ($A.Dist$) and dominance ($D.Dist$).

To conduct analysis of the correlations between phonemes/phonological features against the concreteness/affectiveness, the ratings for each phonemes are calculated as the average of the ratings of the included concepts, grouped by the phonemes and its language, respectively.

Undergoing these procedures, we create a dataset in 142 languages across 21 language families, including ratings in concreteness/affectivness, and phonemes for lemmas. The overall statistics of the data is shown in Table~\ref{tab:dataset}. The map for the data color coded by language families is presented in Fig.~\ref{fig:map}. As shown, the data is highly skewed towards Indo-European languages, and the data is quite scarce in Americas.

\section{Analysis and Results}

\subsection{Colexifications vs. Closeness in Concreteness/Affectivness}

\begin{figure}[ht!]
    \centering
    \includegraphics[width=0.4\textwidth]{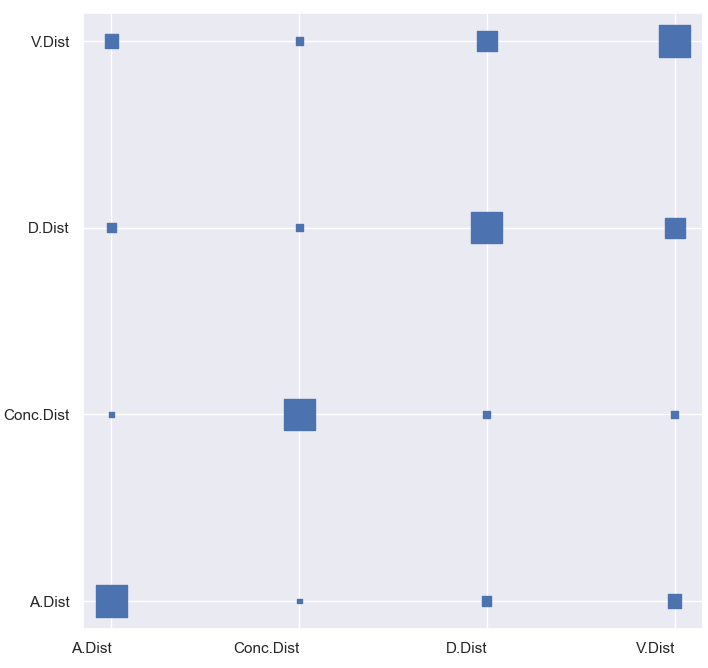}
    \caption{Correlation between Affectiveness- and Concreteness-Distances between the Colexified Concepts. The size of the squares represent correlation coefficients.}
    \label{fig:corr_aff_conc}
\end{figure}

\begin{table}[ht!]
    \centering
     \resizebox{0.6\textwidth}{!}{
    \begin{tabular}{c|cccc}
    \hline 
         &  \textbf{Conc.Dist} & \textbf{V.Dist} & \textbf{A.Dist} & \textbf{D.Dist}  \\\hline
   
     
     \#Colex. & -0.4716* & -0.4192* & -0.5798* & -0.5083*\\
     Colex. Patterns  & -0.4634*  & -0.4115* & -0.581033* & -0.5065*  \\
     
    \#Languages  & -0.4727* & -0.4178* & -0.5798* & -0.5090* \\\hline
    
    \end{tabular}}
    \caption{Correlation between \#Colexifications and the Concreteness/Affectivness Distances between the Colexified Concepts, p-values are in the brackets. The sign * indicates the statistical significance of the correlation at 95\% ($p<0.0001$).}
    \label{tab:corr_colex_aff_conc}
\end{table}

\begin{table*}[htb]
    \centering
    \resizebox{\textwidth}{!}{
    \begin{tabular}{cccccc}
    \hline
         \textbf{Lang. Family} & \textbf{\#Lang.} & \textbf{\# Sample} & \textbf{\# Phonemes}  & \textbf{Initial Phoneme} & \textbf{Last Phoneme} \\\hline
         Turkic & 7 & 2453 & 53 & \textipa{k} (0.1148), \textipa{t} (0.1020) & - \\
         Tai-Kadai & 3 & 2701 & 20 & \textipa{k} (-0.1122), \textipa{n} (0.1066) & -\\
         Austroasiatic & 2 & 3400 &26 & \textipa{\textglotstop} (0.1028) & -\\
         Austronesian & 7 & 21365 & 33 & - & \textipa{\ng} (0.1053)\\
         
         Uralic & 5 & 23352 & 37 & \textipa{V} (-0.1082) & \textipa{i} (0.1423), \textipa{n} (-0.1983), \textipa{6} (0.1005)\\
         Dravidian & 3 & 339 & 22 & \textipa{p} (0.2072) & \textipa{\:l} (-0.2738)\\
         Sino-Tibetan & 5 & 7567 & 39 & \textipa{y} (-0.1189) & \textsuperscript{1} (-0.1428), \textsuperscript{4} (0.1092), \textsuperscript{5} (0.1066) \\
         Afro-Asiatic & 5 & 862 & 44 & \textipa{e} (-0.1450) & \textipa{o} (-0.1107), \textipa{r} (-0.1582),\\ 
          & &  &  &  &  \textipa{B} (-0.1074), \textipa{X} (-0.1432) \\
         \hline 
    \end{tabular}}
    \caption{Correlation between the Initial/Last Phoneme and the Concreteness of Sense Lemma across Languages per Language Family. All the presented coefficients (in the brackets) are statistically significant and at least bigger than 0.1 or smaller than -0.1, corrected with Bonferroni correction ($p<0.05/\#Lang.$*). }
    \label{tab:conc_corr}
\end{table*}

Previous studies show that abstract concepts are often understood by reference to more concrete concepts~\citep{lakoff2008metaphors}, and words that first arise with concrete meanings often later gain an abstract one~\citep{xu2017evolution}.~\citet{XU2020104280} leans on these findings to show that concepts more dissimilar in concreteness and affective valence are more likely to colexify. To test this, we calculate the correlation coefficients\footnote{All the correlation analyses done in this study are using the SciPy implementation of Pearson correlation algorithm. 
} between the number of colexifications and concreteness/affectiveness distances of the colexified concepts across languages. However, the results show the exact contrary to the previous theories and findings. As shown in Table~\ref{tab:corr_colex_aff_conc}, there is a statistically significant and relatively strong negative correlation between colexifications and the distance of concreteness, valence, arousal and dominance. This verifies that it is more likely for a pair of concepts to colexify when they are closer in concreteness and affectiveness.  Our results about affectivness in colexifications is also corroborated by~\citet{di2021colexification}.

Since both distances of conreteness and affectiveness are correlated with colexifications, it is intuitive to assume they might be correlated to each other. To test this, we calcuate the correlation coefficients between each dimension of concreteness and affectiveness. As shown in Fig.~\ref{fig:corr_aff_conc}, the distances of valence and dominance are correlated with each other stronger than other pairs. And, concreteness distance is not significantly correlated with any dimension of affectiveness.

\begin{table}[htb]
    \centering
    \resizebox{0.6\textwidth}{!}{
    \begin{tabular}{c|cccc}
    \hline
        &  Tai-Kadai  &  Austroasiatic  & Indo-European  & Uralic  \\
   \textbf{Features}  & (2822/3) & (3555/2)& (229661/75) & (26795/6) \\\hline
        

    syl     & -0.1570* &  -0.1870* & -0.1851* & -0.2716* \\
        son & -0.1533* &  -0.1698* & -0.1453* & -0.2783*\\
        cons & -0.1734* & -0.2252* & -0.1284* & -0.2092* \\
        cont & -0.1567* & -0.1768* & -0.1520* & -0.2692*\\
        nas & - & -0.1038* & -0.1120* & -0.1718* \\
        voi & -0.1524* & -0.1546* & -0.1726* & -0.2486*\\
        sg & - & -0.1185* & - & - \\
        ant &  -0.1217* & -0.1407* & -0.1553* & -0.2670* \\
        cor &  -0.1574* & -0.1956* & -0.1215* & -0.2195* \\
        distr & - & - & - & -0.1719* \\
        lab & - & - & - & -0.1706*\\
        lo & - & - & - & -0.1244* \\
        
        hi  & -0.1194* & -0.1678* & -0.1015* & -\\
        lo  & -0.1424* & - & - &-\\
        back & -0.1009* & -0.1513* &  - & -\\
        tense & -0.1631* & -0.1175* &  -0.1350* & -0.2675*\\\hline
    \end{tabular}}
    \caption{Correlation between Phonological Features and the Concreteness of Sense Lemma per Language Family. All the presented coefficients are statistically significant and at least bigger than 0.1 or smaller than -0.1, corrected with Bonferroni correction ($p<0.05/\#Lang.$*).}
    \label{tab:corr_feature_vectors}
\end{table}

\begin{table*}[htb]
    \centering
    \resizebox{\textwidth}{!}{
    \begin{tabular}{cccccc}
    \hline
         \textbf{Lang. Family} & \textbf{\#Lang.} & \textbf{\# Sample} & \textbf{\# Phonemes}  & \textbf{Initial Phoneme} & \textbf{Last Phoneme} \\\hline
         Turkic & 7 & 2453 & 53 & \textipa{c} (-0.1178), \textipa{a} (-0.1284) & \textipa{p} (-0.1412), \textipa{y} (-0.1158) \\
         
         Austroasiatic & 2 & 3400 &26 & - & \textipa{h} (-0.1169)\\
         Artificial Language & 2 & 448 & 24 & \textipa{m} (-0.2464)  &  -\\
         
         Dravidian & 3 & 339 & 22 & \textipa{p} (0.1667), \textipa{r} (-0.2044) & \textipa{\:l} (-0.2693) \\
         Sino-Tibetan & 5 & 7567 & 39 & \textipa{p} (-0.1337), \textipa{u} (-0.1272), \textipa{y} (0.1010) & -  \\
         Afro-Asiatic & 5 & 862 & 44 & \textipa{i} (-0.1070), \textipa{j} (0.1065), \textipa{z} (-0.1058), &\textipa{R} (0.1353), \textipa{\textglotstop} (-0.1588) \\
       &  &  &  &  \textipa{g} (-0.1268), \textipa{\textglotstop} (0.1091) & \\\hline 
    \end{tabular}}
    \caption{Correlation between the Initial/Last Phoneme and the Valence of Sense Lemma across Languages per Language Family. All the presented coefficients (in the brackets) are statistically significant and at least bigger than 0.1 or smaller than -0.1, corrected with Bonferroni correction ($p<0.05/\#Lang.$). }
    \label{tab:aff_phoneme}
\end{table*}

\subsection{Phonemes vs. Concreteness/Affectiveness}
Previous studies suggest that characteristics of the initial and the last phoneme have the most significant impact on the phonetic characteristics of the whole phone set~\citep{pimentel-etal-2020-phonotactic}. To test whether there are universals between the initial/last phoneme and the concreteness/affectiveness, we calculate the correlations between them per language family.

Since the whole results are too large to present, we report here only the results where the correlations are statistically significant, and the absolute value of which are bigger than 0.1. To prevent data from incorrectly appearing to be statistically significant, we correct the p-value with Bonferroni correction by dividing it with the number of the languages within the language family that is tested on. Only the results, that are statistically significant at 95\% after applying Bonferroni correction, are reported. 

We can observe that, as in Table~\ref{tab:conc_corr}, by correlating against the concreteness distance, the \textipa{p} as the initial phoneme and the last \textipa{\:l} is significantly and stronger correlated within Dravidian languages, and \textipa{a} in Artificial languages as the first phoneme, compared to others. While across language families, \textipa{k} is correlated with concreteness.

Similarly, we test the correlations against the affectivness distance. Only the results with valence is reported, since the correlations of the phonemes against other affective ratings are not significant. As shown in Table~\ref{tab:aff_phoneme}, \textipa{p} as initials present correlations with affectiveness cross language families, i.e., Sino-Tibetan and Dravidian.

To represent the complexity of phonemes intra language families, we calculate the TTR as the ratio of unique phonemes and the length of all the phonemes for each lemma. Furthermore, the correlation between the TTR and the concreteness/arousal is computed, as shown in Table~\ref{tab:conc_corr}. And also the length of the phoneme segments are calculated for similar correlation test. Across all 8 language families, the segment length is statistically negatively correlated with the concreteness, but positively correlated with arousal. While, the correlations between TTR and the concreteness shows that the more concrete concept, the more diverse (complex) the phonemes are.


\subsection{Phonological Features vs. Concreteness/Affectiveness}

To test whether phonological features of the phonemes correlate with concreteness or affectiveness, for each phoneme/lemma pair, the phonological feature vectors are calculated and the values are aggregated by frequency of the present features. 
As indicated in Table~\ref{tab:corr_feature_vectors}, in the reported data, all the phonological features are negatively correlated with the concreteness. 
While the correlation coefficients in general are quite small, this hints at the possible existence of effects of these phonological features on concreteness.
For instance, the \textit{coronal obstruent (cor)} feature in all four language families is highly negatively correlated with concreteness, indicating that there is a general preference for such words to be abstract in meaning.

\begin{table}[ht!]
    \centering
    \resizebox{0.6\textwidth}{!}{
    \begin{tabular}{ccccc}
    \hline
      \textbf{Lang. Family}   & \textbf{\#Lang.} & \textbf{\# Sample} & \textbf{TTR} & \textbf{LEN}  \\\hline
      \textbf{vs. Concreteness} & & & & \\\hline


       Turkic & 8 & 2557 & - & -0.1373* \\
       Tai-Kadai  & 3 & 2701 & 0.1511* & -0.1834*\\
       Austroasiatic & 2 & 3398 & 0.1794* & -0.2715* \\
       Uralic & 6 & 23508 & 0.1876* & -0.2402*\\
       Dravidian & 3 & 339 &  - & -0.2585* \\
       Indo-European & 75 & 211371 & - & -0.1697* \\
       Sino-Tibetan & 5 & 7567 & 0.1257* & -0.1184*\\\hline 

       \textbf{vs. Arousal} & & & & \\\hline
       Austroasiatic & 2 & 3398 &- &  0.1157* \\
       Mongolic-Khitan & 3 & 66  & - & 0.3294*\\\hline
      
    \end{tabular}
    }
    \caption{Correlation between TTR (Type-to-Token Ratio)/ Segment Length and the Concreteness of Sense Lemma per Language Family. All the presented coefficients (in the brackets) are statistically significant and at least bigger than 0.1 or smaller than -0.1, corrected with Bonferroni correction ($p<0.05/\#Lang.$). }
    \label{tab:my_label}
\end{table}

\section{Conclusion and Future Work}

In this study, we proposed a set of procedures to leverage colexifications to bootstrap cross-lingual datasets, incorporating human ratings of concreteness and affective meanings. The created dataset presents data in 142 languages across 21 language families and 5 language macro areas. However, the procedures can be applied beyond the datasets used in this paper.

Inspired by previous works, we test the correlations between i) the distance of concreteness/affectiveness and the number of colexifications; ii) the phonemes and concreteness/ affectiveness; and  iii) the phonological features and the ratings. It is shown that i) colexifications closer in concreteness/effectiveness are more likely to colexify; ii) certian initial/last phonemes do present statistically significant correlations with the ratings across languages; and iii) there is a positive correlation between the phoneme diversity and concreteness; finally iv) certain phonological features are negatively correlated with the ratings. 
While it is difficult to draw any meaningful conclusions from this finding without a prior hypothesis, we hope that future work can use this dataset to make well-founded findings on the interactions between phonology, concreteness, and affectiveness.

We have showcased the soundness and validity of our approach to curate data from different domains and create a cross-lingual dataset mapping the information. The initial analyses and findings could inspire further applications in NLP and also other fields, such as psychology and psycholingusitics, which we will explore extensively for future work. 

Nevertheless, the analyses conducted in this study are confined to individual correlation tests, which are inadequate for reaching definitive conlusions. For future work, we will employ multivariate modeling techniques utilizing affective/concrete ratings and phonetic features to deepen our understanding of the connections between human conceptualization and sounds across diverse languages and cultures.

\section*{Limitations}

A limitation of this study is the fact that the concreteness ratings of \citet{brysbaert2014concreteness} are curated solely from self-identified US residents. And the affectiveness ratings of~\citet{warriner2013norms} are solely curated in English. As such, there is a risk of an anglocentric bias in the created dataset. 
Nevertheless, the goal of this study is to explore the potential of leveraging colexifications to bootstrap cross-lingual datasets in as many languages as possible, including a lot of low-resource languages.

\section*{Ethics Statement}
Related to the limitations of this work, while this work increases research potential for low-resource languages, this comes with the main ethical risk of potential of propagating the anglocentric bias of some of the source datasets further.

\section*{Acknowledgements}
This work is supported by the Carlsberg Foundation under a \textit{Semper Ardens: Accelerate} career grant held by JB, entitled ``Multilingual Modelling for Resource-Poor Languages'', grant code CF21- 0454. 
We are furthermore grateful to the anonymous SIGMORPHON reviewers for pointing out issues that needed clarification in this work.

\bibliographystyle{unsrtnat}
\bibliography{main}  

\end{document}